\documentclass[10pt,a4paper]{article}

\usepackage{lrec2006}

\usepackage{linguex}
\usepackage{graphicx}
\usepackage{hyperref}

\title{Building a resource for studying translation shifts}

\name{Lea Cyrus}

\address{%
  Arbeitsbereich Linguistik, University of M\"{u}nster\\
  H\"{u}fferstra{\ss}e 27, 48149 M\"{u}nster, Germany\\
  lea@uni-muenster.de%
}

\abstract{This paper describes an interdisciplinary approach which brings
  together the fields of corpus linguistics and translation studies. It
  presents ongoing work on the creation of a corpus resource in which
  translation shifts are explicitly annotated. Translation shifts denote
  departures from formal correspondence between source and target text, i.\,e.
  deviations that have occurred during the translation process. A resource in
  which such shifts are annotated in a systematic way will make it possible to
  study those phenomena that need to be addressed if machine translation
  output is to resemble human translation. The resource described in this
  paper contains English source texts (parliamentary proceedings) and their
  German translations. The shift annotation is based on predicate-argument
  structures and proceeds in two steps: first, predicates and their arguments
  are annotated monolingually in a straightforward manner. Then, the
  corresponding English and German predicates and arguments are aligned with
  each other. Whenever a shift -- mainly grammatical or semantic -- has
  occurred, the alignment is tagged accordingly.}

\begin{document}


\newcommand{\fuse}[0]{FuSe}
\newcommand{\dbliteral}[1]{\texttt{#1}}

\frenchspacing
\maketitleabstract

\section{Introduction}
\label{sec:intro}


Recent years have shown a growing interest in bi- or multilingual linguistic
resources. In particular, parallel corpora (or translation corpora) have
become increasingly popular as a resource for various machine translation
applications. So far, the linguistic annotation of these resources has mostly
been limited to sentence or word alignment, which can be done largely
automatically. However, this type of alignment reveals only a small part of
the relationship that actually exists between a source text and its
translation. In fact, in most cases, straightforward correspondences are the
exception rather than the rule, because translations deviate in many ways from
their originals: they contain numerous \emph{shifts}.

The notion of shift is an important concept in translation studies (see
Section \ref{sec:shifts}). However, shifts have not yet been dealt with
extensively and systematically in corpus linguistics. This paper presents
an ongoing effort to build a resource (FuSe) in which shifts (in translations
from English to German) are annotated explicitly on the basis of
predicate-argument structures, thus making translation equivalence visible.

When finished, the resource will open up a possibility for linguists and
translation theorists to investigate the correspondences and shifts
empirically, but also for researchers in the field of machine translation, who
can use this resource to detect the problems they still have to address if
they want to make their output resemble human translation. The FuSe annotation
project is described in more detail in Section \ref{sec:fuse}, and Section
\ref{sec:related} gives an overview of the way it relates to other work.

\section{Translation Shifts}
\label{sec:shifts}

The investigation of shifts has a long-standing tradition in translation
studies. \newcite{VinayDarbelnet58}, working in the field of comparative
stylistics, developed a system of translation procedures. Some of them
are more or less direct or literal, but some of them are \emph{oblique} and
result in various differences between the source and the target text. These
procedures are called \emph{transposition} (change in word class),
\emph{modulation} (change in semantics), \emph{equivalence} (completely
different translation, e.\,g. proverbs), and \emph{adaptation} (change of
situation due to cultural differences). There is a slight prescriptive
undertone in the work of Vinay and Darbelnet, because they state that oblique
procedures should only be used if a more direct one would lead to a wrong or
awkward translation.  Nevertheless, their approach to translation shifts, even
though \emph{avant la lettre}, continues to be highly influential.

The actual term \emph{shift} was introduced by \newcite{Catford65}, who
distinguishes \emph{formal correspondence}, which exists between source and
target categories that occupy approximately the same place in their respective
systems, and \emph{translational equivalence}, which holds between two
portions of texts that are actually translations of each other. A shift has
occurred if there are ``departures from formal correspondence'' (p. 73)
between source and target text, i.\,e. if translational equivalents are not
formal correspondents. According to Catford, there are two major types of
shifts: \emph{level shifts} and \emph{category shifts}. Level shifts are shifts
between grammar and lexis, e.\,g. the translation of verbal aspect by means of
an adverb or vice versa. Category shifts are further subdivided into structure
shifts (e.\,g. a change in clause structure), class shifts (e.\,g. a change in
word class), unit shifts (e.\,g. translating a phrase with a clause), and
intra-system shifts (e.\,g. a change in number even though the languages have
the same number system). One of the problems with Catford's approach is that
it relies heavily on the structuralist notion of system and thus presupposes
that it is feasible -- or indeed possible -- to determine and compare the
\emph{valeurs} of any two given linguistic items. His account remains
theoretic and, at least to my knowledge, has never been applied to any actual
translations, not even by himself.

The comparative model by \newcite{LeuvenZwart89} has been devised as a
practical method for studying syntactic, semantic, stylistic, and pragmatic
shifts within sentences, clauses, and phrases of literary texts and their
translations.\footnote{There is also a descriptive model, in which the results
  from the comparative model are used to gain insight into shifts on the story
  level and into the norms governing the translation process
  \cite{LeuvenZwart90}. This model is not further discussed, because it is not
  related to the approach presented in this paper.} It consists of four steps.
Firstly, the units to be compared must be established. Van Leuven-Zwart calls
them \emph{transemes}, and they consist of predicates and their arguments or
of predicateless adverbials. Secondly, the common denominator of the source
and the target text transeme -- van Leuven-Zwart calls this the
\emph{architranseme} -- must be determined. In a third step, the relationship
between each transeme and the architranseme -- either synonymic or hyponymic
-- is established. Finally, the two transemes are compared with each other. If
both are synonymic with the architranseme, no shift has occurred. Otherwise,
there are three major categories of shifts: \emph{modulation} (if one transeme
is a synonym and the other a hyponym), \emph{modification} (if both transemes
are hyponymic with respect to the architranseme), and \emph{mutation} (if
there is no relationship between the transemes). There are a number of
subcategories for each type of shift: the whole list comprises 37 items, which
is why the model has sometimes been criticized for being too complex to be
applied consistently.

\section{The FuSe Annotation Project}
\label{sec:fuse}

\subsection{The Data}
\label{sec:data}

The data annotated in FuSe are taken from the Europarl corpus
\cite{Koehn02}\footnote{We use the \textsc{xces} version by
  \newcite{TiedemannNygaard04}.}, which contains proceedings of the European
parliament. In a resource designed for studying translation shifts, it is not
enough that the data be parallel. It is of vital importance that they are
actually translations of each other.\footnote{The Europarl corpus is available
  in eleven languages, so large parts of the English and German subcorpora will
  be translated from a third language.} Since many translation shifts are
directional (e.\,g. \emph{passivisation}), the direction of the translation must
also be clear (in this case from English into German). We used the language
attribute provided by the corpus to extract those sentences that were
originally English. In the corpus, the language attribute is only used if the
language of the corpus file does not correspond with the original
language. Thus, we extracted those sentences from the English subcorpus that
had no language attribute and were aligned to sentences with the language
attribute ``\textsc{en}'' in the German subcorpus. Furthermore, in order to
ensure that the English source sentences were produced by native speakers, we
matched the value of the name attribute against the list of British
and Irish Members of Parliament, which is available on the Europarl
website.\footnote{\url{http://www.europarl.eu.int/}}

\subsection{Predicates and Arguments as Transemes}
\label{sec:pas}

The first step in annotating translation shifts is determining the transemes,
i.\,e. those translation units on which the comparison between source and target
text will be based. It was mentioned in Section \ref{sec:shifts} that the
transemes originally used by Leuven-Zwart (1989) consist of predicates
\emph{and} their arguments (and adverbials). The disadvantage with this
division is that the transemes are quite complex (whole clauses), which means
that there could occur several shifts within one transeme. While this seems to
have been unproblematic for van Leuven-Zwart, who worked with pen and paper,
the units must be smaller in a computational annotation project in order for
the shifts to be assigned unambiguously.

The approach presented in this paper is also based on predicate-argument
structures, because it is assumed that these capture the major share of the
meaning of a sentence and are most likely to be represented in both source and
target sentence. However, unlike in van Leuven-Zwart's approach, each
predicate (lexical verbs, certain nouns and certain adjectives) and each
argument represents a transeme in itself, i.\,e. there are predicate transemes
and argument transemes. Of course, even this more fine-grained annotation
entails that correspondences and shifts on other levels will be missed, but in
order to ensure workability and reproducibility of the annotation, this
restriction seems justifiable.

The predicate-argument structures are annotated monolingually, and since the
annotation is mostly a means to an end, it is kept deliberately simple:
predicates are represented by the capitalised citation form of the lexical
item (e.\,g.\ \textsc{dramatise}).  They are assigned a class based on their
syntactic form (\emph{v}, \emph{n}, \emph{a}, \emph{c}, \emph{l} for `verbal',
`nominal', `adjectival', `copula', and `light verb construction'
respectively). Homonymous predicates are disambiguated for word senses, and
related predicates (e.\,g. a verb and its nominalisation) are assigned to a
common predicate group. In order to facilitate the annotation process, the
arguments are given short intuitive role names (e.\,g.\ 
\textsc{ent\_dramatised}, i.\,e.\ the entity being dramatised). These role
names have to be used consistently only within a predicate group. If, for
example, an argument of the predicate \textsc{dramatise} has been assigned the
role \textsc{ent\_dramatised} and the annotator encounters a comparable role
as an argument to the predicate \textsc{dramatisation}, the same role name for
this argument has to be used. Other than that, no attempt at generalisation
along the lines of semantic cases is made.

If a predicate is realised in a way that might influence the realisation of
its argument structure in a systematic way (e.\,g. infinitive, passive), it
receives a tag to indicate this. If one of the arguments is a relative
pronoun, its antecedent is also annotated. This is done in order to avoid the
annotation of a \emph{pronominalisation} shift (see Section
\ref{sec:gramShift}) in these cases, since the antecedent of relative pronouns
is so close that it would be wrong to call this a pronominalisation. Apart from
this, there is no anaphor resolution. 

\subsection{Shift Annotation}
\label{sec:alignment}

After the predicate-argument structures have been annotated monolingually,
the source predicates and arguments are aligned to their target counterparts.
Sometimes, this is possible in a straightforward manner, like in sentence pair
\ref{ex:noShift}.\footnote{Predicate transemes are in bold face, argument
  transemes are in square brackets. For the sake of clarity, the predicate and
  argument names are omitted.}

\ex.
\label{ex:noShift}
\a. [I\mbox{}] \textbf{refer} [to item 11 on the order of
business].\footnote{Opus/Europarl (en): file ep-00-01-18.xml, sentence id 4.2} 
\label{ex:noShiftEng}
\bg. [Ich] \textbf{beziehe} \textbf{mich} [auf Punkt 11 des Arbeitsplans].\\
I refer me on point 11 of.the workplan.\\
\label{ex:noShiftGer}

However, more often than not the relationship will not be this simple.
Whenever a shift occurs, the alignment between the two predicates or arguments
is tagged. Mainly, the shifts are categorised according to whether they occur
on the level of grammar or on the level of semantics. The following is an
introduction to the main types of shifts. They are first described in Sections
\ref{sec:gramShift} to \ref{sec:problems}, and to make this more concrete, a
few examples are given in Section \ref{sec:ex}

\subsubsection{Grammatical Shifts}
\label{sec:gramShift}

\paragraph{Category Change} This tag is assigned whenever the corresponding
transemes belong to different syntactic categories, and it can be applied
both to predicates and arguments. A typical example would be a verbal
predicate transeme that is translated as a nominal predicate
(nominalisation). 

\paragraph{Passivisation} This tag can only be assigned to the alignment between
verbal predicates (and certain light verb constructions) and is used if an
active predicate has been rendered as a passive predicate.  Often, but not
always, a passivisation shift goes hand in hand with a \emph{deletion} shift
(see below), namely if the source subject is no longer explicitly expressed in
the passivised translation.

\paragraph{Depassivisation} Conversely, if a passive verbal predicate has been
rendered as an active verbal predicate, this is tagged
\emph{depassivisation}. If the source predicate-argument structure lacks the
agentive argument, there will also be an \emph{addition} shift (see below).

\paragraph{Pronominalisation} This tag can only be assigned to
the alignment between arguments. It is used if the source argument is realised
by lexical material (or a proper name) but translated as a pronoun. This tag
is not used if the pronoun in question is a relative pronoun, because the
antecedent can always be found in close vicinity and is annotated as part of
the transeme (see Section \ref{sec:pas}).

\paragraph{Depronominalisation} This tag can only be assigned to the alignment
between arguments. It is used if a source argument transeme is realised as a
pronoun but translated with lexical material or a proper name.

\paragraph{Number Change} This tag is assigned if the corresponding transemes
differ in number, i.\,e. one is singular, the other plural. This happens mainly
between argument transemes, but can also occur between nominal predicates.

\subsubsection{Semantic Shifts}
\label{sec:semShift}

\paragraph{Semantic Modification} This tag is assigned if the two transemes
are not straightforward equivalents of each other because of some type of
semantic divergence, for example a difference in aktionsart between two verbal
predicates. 

It is rather difficult to find objective criteria for this shift. In the
majority of cases two corresponding transemes exhibit some kind of divergence
if taken out of their context, but are more or less inconspicuous
translations in the concrete sentence pair. Since an inflationary use of this
tag would decrease its expressiveness, semantic likeness is interpreted
somewhat liberally and the tag is assigned only if the semantic difference is
significant. Of course, this is far from being a proper operationalisation, and
we hope to clarify the concept as we go along.

\paragraph{Explicitation} This is a subcategory of \emph{semantic
  modification}, which is assigned if the target transeme is lexically more
specific than the source transeme. A clear case of explicitation is when
extra information has been added to the transeme. One could also speak of
explicitation when a transeme has been depronominalised (see Section
\ref{sec:gramShift}). However, since the \emph{depronominalisation} shift is
already used in these cases, this would be redundant and is therefore not
annotated.

\paragraph{Generalisation} This is the counterpart to the \emph{explicitation}
  shift and is used when the target transeme is lexically less specific than
  its source, and in particular if some information has been left out in the
  translation. To avoid redundancy, it is not used for
  \emph{pronominalisation} shifts. 

\paragraph{Addition} This tag is assigned to a target transeme, either
predicate or argument, that has been added in the translation process. For
instance, if there has been a \emph{depassivisation} shift and if the agentive
argument had not been realised in the source text, it must be added in the
target text. Note that we do not speak of addition if only \emph{part} of the
transeme has been added. In this case, the \emph{explicitation} tag is to be
assigned (see above).

\paragraph{Deletion} This tag is assigned to a source transeme that is
untranslated in the target version of the text. Analogous to the
\emph{addition} shift, this tag is only used if the entire transeme has been
deleted. If it is only part of a transeme that is untranslated, the shift is
classified as \emph{generalisation}. 

\paragraph{Mutation} This tag is used if it is possible to tell that two
transemes are translation equivalents (in the sense intended by Catford, see
Section \ref{sec:shifts}), but if they differ radically in their lexical
meaning. This shift often involves a number of other shifts
as well.

\subsubsection{Problematic Issues} 
\label{sec:problems}

\paragraph{Long Transemes} Normally, a maximum of two shifts can be assigned
to any one pair of transemes: a grammatical and a semantic shift. This can be
a problem if the transemes are long, like for instance clausal
arguments. Because of their length, they can contain multiple shifts, and it is
difficult to determine which of them is to be the basis for the shift
annotation, in particular if they are contradictory (e.\,g. there might occur
both \emph{generalisation} and \emph{explicitation} in different parts of the
transeme). The general rule here is to check whether the shift actually
affects the overall transeme. In most cases, long transemes contain further
transemes, e.\,g. clausal arguments contain at least one extra predicate plus
arguments, which will be represented by their own predicate-argument
structure, and it is on this level that these shifts are recorded.

\paragraph{Lexical Modals} Modal auxiliaries are currently not annotated as
separate predicates. This is no problem as long as the modality is expressed
by means of a modal auxiliary in both languages. However, sometimes modality
is expressed by a full verb with modal meaning (e.\,g. \emph{to wish}), which is
consequently annotated as a predicate. If the other language uses a modal
auxiliary, no alignment is possible, because there is no predicate transeme.
Normally, when a predicate transeme has no correspondent in the other
language, one would assign the \emph{addition} or \emph{deletion} shift, but
since nothing really has been added or deleted, this is not a particularly
satisfying solution.  One way out would be to rethink our attitude towards
modals and simply annotate them as predicates.  While the decision is still
pending, such predicates are tagged \emph{dangling modal}.

\paragraph{Structure Shifts} It also happens that a transeme cannot be
aligned because it is not realised as part of a predicate-argument structure
in the other language. An example of this would be a full verb with modal
meaning that is rendered as an adverb in the other language (e.\,g. \emph{to
  wish} -- \emph{gern}, `with pleasure').  Again, it would not be adequate to
speak of addition or deletion. However, since these cases constitute real
structural shifts, the additional tag \emph{non-pas} (i.\,e.
`non-predicate-argument-structure') has been introduced to deal with them.

\subsubsection{Examples}
\label{sec:ex}

In this section, the shift annotation described in the previous sections is
illustrated by a few examples from the corpus.

\ex. 
\label{ex:dramatise}
\a. [It] should not be \textbf{dramatised} [into something more than
that].\footnote{Opus/Europarl (en): file ep-00-01-18.xml, sentence id 8.4}
\label{ex:dramatiseEng}
\bg. [Wir] sollten [die ganze Sache] nicht weiter \textbf{aufbauschen}.\\
We should the whole thing not further exaggerate.\\
\label{ex:dramatiseGer}
 
Both sentences contain one predicate transeme (\textsc{dramatise} and
\textsc{aufbauschen}) and two argument transemes. The two predicates differ
with respect to voice: while the source predicate in \ref{ex:dramatiseEng} is
passive, its German counterpart \ref{ex:dramatiseGer} is active, so the
alignment between these two predicates would receive the
\emph{depassivisation} tag. As a consequence of the change of voice, the
agentive argument, which is left unexpressed in the passive source sentence,
is explicitly expressed in the German translation (\emph{Wir}, `we'), and is
consequently tagged \emph{addition}.  Conversely, the argument \emph{into more
  than that} is left unexpressed in the German version -- this is an instance
of \emph{deletion}.  Furthermore, the subject of the English sentence
(\emph{it}), the entity that is being dramatised, is expressed lexically in
the translation. The alignment between these two arguments is thus tagged as
\emph{depronominalisation}.

\ex.
\label{ex:aktionsart}
\a. [\dots] [we] agreed yesterday to \textbf{have} [the Bourlanges report] [on
today's agenda].\footnote{Opus/Europarl (en): file ep-00-01-18.xml, sentence
  id 11.1 (abbreviated for convenience).}
\label{ex:aktionsartEng}
\bg. [Wir] kamen gestern \"{u}berein, [den Bericht Bourlanges] [auf die
Tagesordnung von heute] zu \textbf{setzen}.\\ 
We came yesterday agreed, the report Bourlanges on the agenda of today to
put.\\
\label{ex:aktionsartGer}

In this sentence pair, the alignment between the two predicate transemes
\textsc{have} and \textsc{setzen} is tagged \emph{semantic modification}
because they differ in aktionsart: the English predicate is static, while the
German predicate is telic.

\ex.
\label{ex:gen}
\a. [I\mbox{}] do not want to \textbf{drag up} [the issue of this building]
endlessly [\dots]\footnote{Opus/Europarl (en): file ep-00-01-18.xml, sentence
  id 13.3 (abbreviated for convenience)}
\label{ex:genEng}
\bg. [Ich] will nicht endlos [auf diesem Thema] \textbf{herumreiten} [\dots]\\
I want not endlessly on this topic keep.on.about [\dots] \\
\label{ex:genGer}

Example \ref{ex:gen} illustrates the use of the \emph{generalisation} shift.
The second argument transeme in the original \ref{ex:genEng} contains explicit
information on what the issue is about. This information is left out in the
translation \ref{ex:genGer}, with the result that the transeme is more general.
Since it is only a part of the transeme that has been dropped in the
translation, this is not annotated as deletion.

\subsection{Tools}
\label{sec:tools}

\newcommand{\fuser}[0]{FuSer}
\newcommand{\annotate}[0]{\textsc{Annotate}}

The (monolingual) predicate-argument structures are annotated with \fuser\ 
\cite{PykaSchwall06}.  The annotator is presented with a sentence and, if
available,\footnote{The original outline of FuSe also included phrase
  structure \cite{CyrusEtAl04,CyrusFeddes04}, but this was shelved for
  practical reasons.  However, syntactic annotation is something from which
  FuSe would definitely profit, and the tools can be used both on raw data and
  on trees.} a graphical view of its syntactic structure, and selects those
tokens (or nodes from the tree) which are to be annotated as a predicate. The
annotator can choose from a list of predicates, or, if the predicate type is
encountered for the first time, add a new predicate type or group to the
database.  Once the predicate is annotated, the procedure is repeated for the
arguments of this predicate. Again, either the argument types are chosen from
the list or added to the database. Additionally, the necessary tags (see
Section \ref{sec:pas}) are added to the predicates and arguments. The
annotation process is then repeated for all the predicate-argument structures
in a sentence. They are annotated independently, i.e. there is no nesting of
predicates. 

Currently, the predicate-argument structures are annotated manually, which is
a time-consuming task. However, there are a couple of ``wizards'' under
development which will assist the annotator. For instance, there will be a
wizard to scan the sentence for predicate candidates or to suggest suitable
argument types when the predicate is already included in the database.

Technically, \fuser\ is a platform-independent Java application which operates
on an extended \annotate\ MySQL database. This data model makes it possible to
be flexible with respect to the input data, which can be either raw (as is
currently the case) or syntactically annotated. Furthermore, since the
\annotate\ database is only extended and not modified, data processed with
\fuser\ can always be processed by \annotate\ afterwards (e.\,g. for further
annotation).

\begin{figure*}[htbp]
  \centering
  \includegraphics[width=\textwidth]{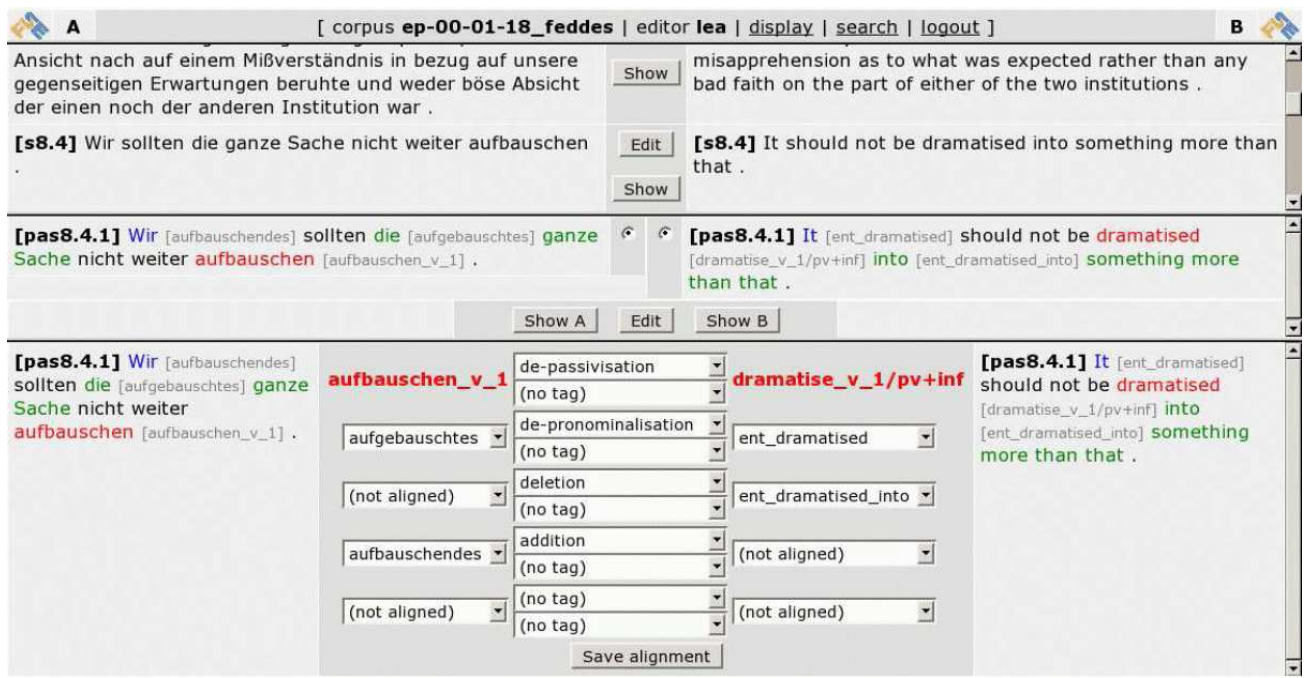}
  \caption{Screenshot of the Web-based alignment tool, showing the annotation
  of Example \ref{ex:dramatise}}
  \label{fig:sshot}
\end{figure*}

It is planned to extend \fuser\ for the bilingual alignment and the shift
annotation. While this extension is under development, we use a simple
Web-based alignment tool (XML, Perl, CGI) for this task (see Figure
\ref{fig:sshot}). The browser window is divided into three parts: in the upper
third, the annotator can select a sentence pair. In the middle part, all the
predicate-argument structures that have been annotated for these sentences are
listed, with the predicates and arguments being highlighted in different
colours. The annotator chooses (by means of radio buttons) two corresponding
predicate-argument structures, which are then displayed in more detail in the
lower window. Here, the annotator can align corresponding predicates and
arguments with each other and, if necessary, choose up to two shift-tags for
each pair of transemes from a drop-down menu. The lower window can also be
used for viewing existing annotation.

\section{Related Work}
\label{sec:related}

Being interdisciplinary, this work is related both to approaches
from translation studies and to various annotation projects. Since the
translational approaches have already been presented in Section \ref{sec:shifts},
this section will confine itself to related annotation projects and the way
they compare with FuSe. 

First of all, there are those projects that also deal with predicate-argument
structures in some way, in particular FrameNet \cite{RuppenhoferEtAl05} (which
is mainly a lexicographical project but can, of course, be adopted for
extensive corpus annotation, as is currently done in the \textsc{salsa}
project \cite{ErkEtAl03a}), PropBank \cite{PalmerEtAl05}, and NomBank
\cite{MeyersEtAl04b}. In these projects, the predicate-argument annotation is
the main objective, so they all try some kind of generalisation by organising
their predicates in semantic frames (FrameNet) or by following the Levin classes
(PropBank, and for nominalisations also NomBank). In FuSe, however, this type
of annotation is not an end in itself -- predicates and their arguments
simply constitute the transemes. Consequently, their annotation is kept
deliberately simple and is entirely predicate-group specific without any
attempt at generalisation. 

What distinguishes FuSe from all of the above mentioned projects is that it
deals not with one language, but with two (and potentially more) languages,
and in particular with parallel data. It thus makes sense to also compare it
with approaches that model the relationship between original texts and their
translations.  

In the \textsc{iamtc} project \cite{FarwellEtAl04}, texts from six languages
(Arabic, French, Hindi, Japanese, Korean, and Spanish) and their translations
into English are annotated for interlingual content. For each original text,
at least two English translations are being annotated (so as to be able to
study paraphrases), and the annotation proceeds incrementally over three
increasingly abstract levels of representation.

The difference here, apart from the languages involved, lies first and
foremost in the type of the semantic representation. The semantic
representation aimed at in \textsc{iamtc} will be a full-fledged interlingua
and thus far more complex than the predicate-argument structure in FuSe.
The ultimate aim is to create a full semantic representation of each sentence
that is not only independent of the actual syntactic realisation, but also of
the language. Thus, provided there aren't any shifts, the \textsc{iamtc}
representations of the source and target language material could be identical.

In FuSe, however, considerable parts of the sentence meaning are not captured
by the predicate-argument annotation. Furthermore, the annotation is entirely
language-specific. There is nothing in the database that indicates that a
predicate \textsc{buy} and a predicate \textsc{kaufen} can be used to express
the same meaning in their respective languages, except for the fact that they are
being aligned with each other. The predicate-argument structure is the basis
of the alignment, but it is not an interlingua.   

Furthermore, in \textsc{iamtc}, there seems to be no direct alignment between
the different versions of the texts. Differences in semantics result in
differences in the interlingual representation, but particularly shifts on
the level of grammar, e.\,g. passivisation, are normalised even on the most
basic level (cf. p. 58).

As part of the Nordic Treebank
Network,\footnote{\url{http://w3.msi.vxu.se/~nivre/research/nt.html}}
\newcite{VolkEtAl06} have begun to build an English-Swedish-German treebank in
which the relationship between the languages is annotated by alignment on a
sub-sentential level, i.\,e. between words, phrases, and clauses. In this
respect, there is a close resemblance with FuSe. One of the differences is
that their emphasis lies on the syntactic annotation of the sentences, which
is not the case in FuSe.

Second, the phrase alignment is done directly, i.\,e. without the
predicate-argument ``detour'': nodes that ``convey the same meaning and could
serve as translation units'' are aligned, and there are two types of alignment,
namely \emph{exact} and \emph{approximate}. 

\section{Outlook}
\label{sec:outlook}

So far, the annotated data consist of English source texts that have been
translated into German. It would be interesting to include the opposite
direction as well, i.\,e. German source texts that have been translated into
English. This would make it possible -- by comparing the types of shifts and
their quantity -- to find out which shifts have occurred due to the direction
of the translation process, and which shifts might be due to the translation
process as such (e.\,g. \emph{explicitation} is taken to be such a potential
``translation universal'' in current translation research, see
\newcite{MauranenKujamaeki04}).

Furthermore, the genre of the Europarl corpus -- parliamentary proceedings
-- is highly restricted and it would be a useful extension to include other
types of data (e.\,g. technical language, literary prose) in order to compare
the occurrence of shifts across genres.

\section{Acknowledgements}
\label{sec:thanks}

I would like to thank Hendrik Feddes, Robert Memering, Frank Schumacher, and
the three anonymous reviewers for helpful and valuable comments.


\bibliographystyle{lrec2006}
\bibliography{lrec06}



\end{document}